%% file: main.tex
\documentclass[11pt]{article}
\usepackage[margin=1in]{geometry}
\usepackage{amsmath,amssymb}
\usepackage{booktabs}
\usepackage{graphicx}
\usepackage{float}
\usepackage{longtable}
\usepackage{microtype}
\usepackage[hidelinks]{hyperref}

\newcommand{\code}[1]{\texttt{#1}}
\newcommand{\Grid}{\textsc{Grid-crop}}
\newcommand{\SLMc}{\textsc{StaticLM-crop}}
\newcommand{\FLMc}{\textsc{FitLM-crop}}
\newcommand{\SLMr}{\textsc{StaticLM-rot}}

\title{Following the Error Through a Compressed Network:\\
       An Angular Measure for FFN Weight Compression in ViT-L/16}
\author{Paolo D'Alberto\thanks{With the full collaboration of Codex (Claude Sonnet 5.5), an AI agent that ran the experiments on the GPUs,
        wrote the scripts, the notes and this draft under the author's direction. Every number in the text is
        taken from a script and its result file.}\\
        Advanced Micro Devices, Inc.\\
        \texttt{paolo.dalberto@amd.com}}
\date{}

\begin{document}
\maketitle

\input{sections/00_abstract}

\input{sections/01_introduction}

\input{sections/02_related}

\input{sections/03_notation}

\input{sections/04_schemes}

\input{sections/04b_span_inputs}

\input{sections/05_observations}

\input{sections/06_results}

\input{sections/07_conclusion}

\input{sections/08_appendix}

\bibliographystyle{plain}
\bibliography{refs}
\end{document}

%% file: sections/00_abstract.tex
\begin{abstract}
We study the compression of the 48 feed-forward network (FFN) matrices
of the Vision Transformer ViT-L/16 to 2--8 bits with four schemes: a
per-token grid, a fixed Gaussian Lloyd--Max table, a table fitted to
the matrix, and a random-sign Hadamard rotation followed by the fixed
table. We evaluate on 50,000 images. At 4 bits the top-1 accuracy of
the four schemes lies within 0.8 points of the float32 model, while
the agreement with its predictions ranges from 98.4\% to 94.4\%.
We use the angular error between the compressed and the float32
residual stream, followed through the blocks, as a portable measure that
ranks the schemes as the final classification does and shows where
the ranking is built: with rotation the angle grows close to the
square root of the number of blocks, the signature of independently accumulating
errors, while the other schemes are amplified in the later blocks.
The same measure shows that a scheme that is better on one block in
isolation can be worse in the whole network. To address this
inconsistency we introduce an angular measure of the spectrum of the output of a block, which in isolation orders the schemes as the whole network does where they diverge most.
We support these observations with real samples and with statistically generated ones.
The scope is one model, the FFN weights only, and float evaluation of
compressed weights; no speed-up is claimed. The error measures and the
sample generation, however, can be applied to other models and precisions.
\end{abstract}

%% file: sections/01_introduction.tex
\section{Introduction}
\label{sec:intro}

The feed-forward network (FFN) matrices hold most of the parameters of a Vision Transformer (ViT). In ViT-L/16 the 48 matrices
($\mathrm{fc1}\colon 1024\to4096$ and $\mathrm{fc2}\colon 4096\to1024$ in each of 24 blocks) hold 201 of its 304 million parameters.
Compressing them to a few bits per weight reduces storage and traffic, and the question this paper asks is how to \emph{compare}
compression schemes, not which one to build.

Three habits are common, and each can mislead.
The first compares the error made on the weights, how far each compressed weight vector is from the original. On this model a table fitted to the matrix is closer to the weights than a plain per-token grid, yet the compressed network is closer to the original with the grid.
The second compares top-1 accuracy on a sample. At 4 bits, on 50,000 validation images, four schemes differ from the float32 model by
0.15 to 0.78 points of top-1, but behind this small gap in the total hides a large movement of labels: 802 to 2,792 images change their predicted class, some are negative flips (right before, wrong now, which we call \emph{broken}) and some positive flips (wrong before, right now, \emph{fixed}), and the two nearly cancel in the accuracy. On a sample of 5,000 images the order of
their top-1 is the reverse of the order of the damage.
The third compresses one block at a time and keeps the scheme that is best there. In our data the preference measured on one block
does not survive when all blocks are compressed.

We use instead measures that each answer one intuitive question. The \emph{angle} between the compressed and the float32 residual stream after each block, per token, says how far compression has turned the token, that is, the embedding, away from where it should point; since a transformer reads direction much more than length, a turn is the damage that matters, and turns of independent errors add up in a way we can predict. The \emph{agreement} with the float32 prediction says how often the compressed network still says the same thing as the original, right or wrong. The Kullback--Leibler (KL) divergence of the output distribution says how much the network's confidence has moved, even when its answer has not. The \emph{swing} says who pays for a small change in accuracy: the images whose class changes, negative flips (broken: right before, wrong now) and positive flips (fixed: wrong before, right now), which can cancel in the accuracy and not in the damage. Inside one matrix, the \emph{span} of the output (the leading eigenvectors of its second-moment matrix, where the output keeps its energy) says whether compression has kept the eigenvectors that carry the signal, or has bent or shrunk them; losing a direction is the damage that no later block can repair. The same angular idea measures the weights (Section~\ref{sec:notation}), so one kind of quantity connects the weight, the stream and the output.

\paragraph{Contributions.}
(i) A measure and a protocol: the stream angle followed through the blocks, evaluated on named samples of 100 to 50,000 images with paired intervals, and the swing of the classification, its negative and positive flips.
(ii) The error flow: with rotation the stream angle grows as the square root of the number of blocks at every width from 8 to 2 bits (exponents 0.43 to 0.53), the signature of independent errors, and the crop schemes rise above this line in the later blocks.
(iii) Isolation and context: the error that arrives at a block is four to five times what the block adds, the order of two schemes measured on single blocks reverses in the whole network, and the gap between them arises in fc2.
(iv) The span: a spectrum angle and other measures of the output of a matrix show where a scheme loses energy, and order the schemes in isolation as the whole network does where they diverge most.
(v) Generated inputs: from 500 images a bulk and one hot eigenvector reproduce the span measures of the real inputs on held-out images.
(vi) A sweep of four schemes at 8, 6, 5, 4, 3 and 2 bits, and a confirmation on 50,000 images at 4 bits.

\paragraph{Scope.}
One model (ViT-L/16, the \code{augreg\_in21k\_ft\_in1k} checkpoint), the FFN weights only, one random sign seed for the rotation, and
float32 evaluation: the compressed weights are decoded to float before the multiply, so we measure the effect of the representation, not
of a kernel. We report no run time. The rotation costs a Hadamard transform of the activations at each FFN matrix, which we do not
measure here.

Section~\ref{sec:related} places the work, Section~\ref{sec:notation} defines the objects and measures, Section~\ref{sec:schemes}
the schemes, Section~\ref{sec:span} the span, as an error measure in context and as a generator of tokens, Section~\ref{sec:obs} states the observations as testable statements, and Section~\ref{sec:results} gives the results.
Section~\ref{sec:conclusion} lists the limits.

%% file: sections/02_related.tex
\section{Related work}
\label{sec:related}

\emph{Rotation before rounding.} Multiplying the weights and the activations by an orthogonal matrix before quantization spreads
outliers over the coordinates; it is used with randomized Hadamard transforms or learned rotations in recent work on language models
\cite{quip,quarot,spinquant}. We use a fixed random-sign Hadamard rotation and do not propose a new one; we ask what the rotation does to the
error through the blocks of the network.
\emph{Data-aware weight quantization.} GPTQ and AWQ use calibration data to choose the rounding or the scales \cite{gptq,awq}. Our schemes use
no data (information levels 0 to 2, Section~\ref{sec:schemes}); data would be a further level.
\emph{Non-uniform formats.} Lloyd--Max quantizers \cite{lloyd,max} minimize the mean squared error for a density; fixed non-uniform tables are
used in 4-bit formats for language-model weights \cite{qlora}. We use the same construction for a Gaussian table solved once and for a table fitted to a matrix.
\emph{Models.} The network is a vision transformer \cite{vit} in the augmented-regularized training recipe \cite{augreg}.

The contribution here is not a quantizer. It is the measure (the angle followed through the network), the protocol (paired intervals on named samples,
the swing), and what they show about isolation, accumulation and localization. The angular error and the statistical treatment of the compression error, with ablation and validation on samples, were developed for the compression of the key--value cache
\cite{ablation,tqinference}; here we apply the same machinery to the weights of a public model, and extend it from one layer to the error that flows through the whole network.

%% file: sections/03_notation.tex
\section{Notation and definitions}
\label{sec:notation}

We fix the vocabulary used in the rest of the paper: the block and its parts, what we call an outlier, the two kinds of angle, the classification measures, and the samples on which they are computed.

A block $b\in\{0,\dots,23\}$ of ViT-L/16 maps the residual stream $x_b$ to $x_{b+1}$:
\begin{align}
x'_b &= x_b + \mathrm{Attn}\big(\mathrm{LN}(x_b;\gamma_1,\beta_1)\big),\\
x_{b+1} &= x'_b + W_2\,\mathrm{GELU}\big(W_1\,\mathrm{LN}(x'_b;\gamma_2,\beta_2)+b_1\big)+b_2,
\end{align}
where $\mathrm{LN}(x;\gamma,\beta)=\gamma\odot(x-\mu)/\sigma+\beta$ is layer normalization, with $\mu$ and $\sigma$ the mean and the standard deviation of the 1024 entries of the token $x$ and with a learned scale $\gamma$ and shift $\beta$; the two layer normalizations of a block are the same operation with their own parameters $(\gamma_1,\beta_1)$ and $(\gamma_2,\beta_2)$, and every block has its own. $\mathrm{GELU}$ is the Gaussian error linear unit, a smooth activation that suppresses most negative values, and with $W_1\in\mathbb{R}^{4096\times1024}$ (fc1) and $W_2\in\mathbb{R}^{1024\times4096}$ (fc2), stored as [out, in]. Only $W_1$ and $W_2$ are compressed;
attention, embeddings, norms and the head stay in float32.

Two kinds of token appear. The network cuts an image into $14\times14=196$ patches of $16\times16$ pixels and adds one class token; each of these 197 positions is a vector of 1,024 numbers that the blocks update, and we call it an \emph{activation token}.
A \emph{weight token} is one row $w_j\in\mathbb{R}^n$ of a weight matrix, that is, the weight vector of one output channel ($n=1024$ for fc1, $4096$ for fc2).
We place a weight token on the sphere, $\hat z_j=\sqrt n\,w_j/\|w_j\|$, so that its root mean square is 1; we call $\sigma_j=\|w_j\|/\sqrt n$ the sigma of the token (a root mean square, with no mean subtracted), measure coordinates in sigma units, $|w_{ji}|/\sigma_j$, and store the norm separately.

An \emph{outlier} is a coordinate of a weight token whose magnitude (its absolute value) in sigma units is far larger than that of the other coordinates of the token. Counting the coordinates beyond a threshold is one measure
(Section~\ref{sec:results}); the largest coordinates of the 48 matrices reach 10 to 37 sigma, where the largest coordinate of a Gaussian matrix of the same shape would reach about 5.5.
By extension we also call a weight token that contains outlier coordinates an \emph{outlier token} (in the results, a heavy row), and the positions of those coordinates its outlier columns.
We use a second, functional definition: an outlier is a coordinate that spends bits on range rather than on precision. A rotation does not remove it; it spreads its energy over
all coordinates, so that fewer bits cover the range and more coordinates carry the same information.

The angle between two vectors $u$ and $v$ is $\theta(u,v)=\arccos\big(\langle u,v\rangle/\|u\|\|v\|\big)$. For a weight token $w$ and its compressed copy $\tilde w$ the \emph{weight angle} is $\theta(w,\tilde w)$.
The weight angle measures how far compression has turned the token. Any orthogonal transformation, such as a rotation, preserves angles.

Take the first block and a token $t$ of its input $x_0$. The block turns $t$ into the token $r$ of $x_1$; the compressed block turns the same $t$ into $r'$. The \emph{stream angle} is $\varphi_0(t)=\theta(r,r')$: how far compression has turned the output token.
Each token is first centred, that is, the mean of its 1,024 numbers is subtracted. We do this because layer normalization is invariant to a constant added to all the numbers of a token: it subtracts the mean itself, so the next layer cannot see such a shift.
For a later block $b$ the compressed network has also compressed the blocks before it, so $r$ and $r'$ are the tokens of $x_{b+1}$ in the float32 and in the compressed network, and $\varphi_b(t)$ contains the error accumulated so far.
We report the median (and the 99th percentile) over tokens.

If every block adds an independent error of similar size, the errors add like random steps, in quadrature, so after $n=b+1$ compressed blocks the median angle grows as $n^{1/2}$ for small angles.
We fit $\varphi_n\propto n^{p}$ and call $p$ the \emph{growth exponent}: $p$ near $1/2$ means independent accumulation, and a larger $p$ means that later blocks amplify the error.

For an image, the \emph{logit angle} is $\theta(l,l')$, where $l$ and $l'$ are the 1,000 logits of the float32 and of the compressed network, each centred by removing a constant, its mean; the softmax cannot see such a constant. We could compare the probabilities directly, with a cross entropy (we also report the Kullback--Leibler divergence of the output), but the angle is a measure common to the weights, the stream and the logits.

\emph{Agreement} is the share of images with the same predicted class as the float32 network. \emph{Top-1} and \emph{top-5} are the usual accuracies.
\emph{KL} is the divergence of the compressed softmax from the float32 one. The \emph{swing} counts the images whose class \emph{changed}, among them those the float32
network classified correctly and the compressed one does not (a \emph{negative flip}, which we also call \emph{broken}), and the reverse (a \emph{positive flip}, \emph{fixed}) \cite{pct}; the two counts are the discordant pairs of McNemar's test \cite{mcnemar}. The top-1 change is fixed minus broken, over the sample;
it can be near zero while many images change.

The validation images are taken in file order. A sample is named by its number of images and the number of classes it covers (Table~\ref{tab:samples}).

\begin{table}[H]
\centering\small
\begin{tabular}{llrrp{5.2cm}}
\toprule
sample & selection & images & classes covered & used for\\
\midrule
100-93 & every 500th image & 100 & 93 & checking the scaffolding\\
500-401 & every 100th image & 500 & 401 & per-token flow runs, swap tests, local analysis\\
5000-995 & every 10th image & 5,000 & 995 & exploration\\
50000-1000 & all images & 50,000 & 1,000 & confirmation\\
\bottomrule
\end{tabular}
\caption{The samples. Only the full validation set is balanced across classes.}
\label{tab:samples}
\end{table}

The float32 network has top-1 85.84\% and top-5 97.82\% on 50000-1000 (85.86\% top-1 on 5000-995).
All evaluation uses batch size 10; the noise floor of the logits between runs is about $3\times10^{-6}$.
Intervals are 95\% bootstrap intervals over images (2,000 resamples). The same resamples are used for every arm, so a difference between two schemes is a paired difference.

%% file: sections/04_schemes.tex
\section{Compression schemes and their assumptions}
\label{sec:schemes}

We study the effect of four compressions of a weight token to $b$ bits per coordinate: a uniform grid, as in an integer representation; Lloyd--Max tables, which minimize the mean squared error (one fixed table for Gaussian coordinates and one fitted to the matrix); and a Hadamard rotation, which spreads the outliers over all coordinates, combined with the fixed table.
In each, a coordinate is replaced by the index of one of $2^b$ levels and the token keeps one scale chosen by least squares.
We rank them by what the encoder must know about the weights, the \emph{information level}: level 0 nothing, level 1 the token itself, level 2 the distribution of the whole matrix, and level 3 data, such as calibration inputs of the layer, whose covariance an encoder could use. We stay at levels 0 to 2.

\paragraph{Crop with a grid (\Grid).}
A weight token is clipped to an interval $[-c\sigma,c\sigma]$, with $\sigma$ its sigma (the root mean square), that is, to $[-c,c]$ on the sphere of radius $\sqrt n$ where the token has sigma 1; the grid is the $2^b$ equally spaced levels of an integer representation, one of which is zero, with the largest level at $c$; each coordinate takes the nearest level.
The interval is chosen per token, from candidates of 1 sigma upward, as the one with the smallest weight angle, and the token keeps the least-squares scale.
It assumes nothing about the distribution of the coordinates and needs the token itself (information level 1).

\paragraph{Lloyd--Max tables.}
A Lloyd--Max quantizer chooses levels and thresholds that minimize the mean squared error under a density; the levels are the conditional means of the cells \cite{lloyd,max}.
\SLMc{} uses the table of the standard normal, solved once for each width, on the normalized token: it assumes Gaussian coordinates and needs nothing about the weights (level 0).
\FLMc{} fits the table to the pooled coordinates of the matrix itself (level 2): it replaces the Gaussian assumption by the empirical density, at the price of using the weights to build the table.
The fit iterates until the relative decrease of the distortion is below $10^{-9}$, with a cap of 10,000 iterations.

\paragraph{Rotation.}
Let $H$ be the $n\times n$ Walsh--Hadamard matrix scaled by $1/\sqrt n$ and $d$ a vector of random signs; the transform $Tx=H(d\circ x)$, where $\circ$ is the element-wise product, is orthogonal. Since $y=Wx=(WT^{-1})(Tx)$, we compress the rows of $WT^{-1}$ and apply $T$ to the activation $x$ at run time. We expect a random rotation to send a token to a point close to uniform on the sphere; its coordinates then follow a beta distribution (scaled) and, as $n$ grows, a Gaussian. The random-sign Hadamard transform is a cheap approximation of such a rotation, and we test the Gaussian expectation (Section~\ref{sec:results}).
The cost is a transform of length $n$ per activation token ($n\log_2 n$ additions for $n=1024$ or $4096$), and the sign vector.

\paragraph{Rotation with the fixed table (\SLMr).}
We rotate the weight token with $T^{-1}$ and quantize it with the \SLMc{} table.

\begin{table}[H]
\centering\small
\begin{tabular}{p{2.8cm}p{5.4cm}p{4.8cm}}
\toprule
scheme & quantizer & information about the weights\\
\midrule
\Grid & uniform grid, per-token clip & the token (level 1)\\
\SLMc & $N(0,1)$ Lloyd--Max & none (level 0)\\
\FLMc & fitted Lloyd--Max & the matrix (level 2)\\
\SLMr & $N(0,1)$ Lloyd--Max after rotation & none (level 0)\\
\bottomrule
\end{tabular}
\caption{The four schemes. Information levels: 0 nothing about the weights, 1 the token, 2 the distribution of the matrix, 3 data.}
\label{tab:schemes}
\end{table}

The bit width counts the indices only; the per-token scale (and, for the grid, the clip) is side information of the order of $1/64$ bit per weight if stored in 16 bits per token of 1,024 weights.

%% file: sections/04b_span_inputs.tex
\section{The span: an error measure in context and a generator of tokens}
\label{sec:span}

To see how the coordinates of a set of tokens relate to each other we look at their empirical covariance, a matrix; if the coordinates were independent Gaussians it would be diagonal. We keep the mean in it, which makes it a second moment: the energy of the tokens.
An eigenvalue and eigenvector analysis of this matrix shows the eigenvectors that carry the energy and, by their eigenvalues, how much each carries; the \emph{span} is the subspace of the leading eigenvectors, and its dimension is the number of directions that matter. We use it in two ways.
First, as an error measure that depends on the context: the output of a matrix lives in a span, and compression can shrink it, tilt it or lose part of it; it measures the error where the inputs put the energy (Section~\ref{sec:spanmeasures}, Table~\ref{tab:measures}).
Second, to generate tokens: the covariance describes how the tokens of a small sample are related to each other, and a description of a few numbers lets us generate new tokens that keep what matters, so that a small sample generalizes (Section~\ref{sec:smallsample}).

\subsection{The span as an error measure}
\label{sec:spanmeasures}
Take fc2 of one block, $W\in\mathbb{R}^{1024\times4096}$, and its compressed copy $\tilde W$. Let $A\in\mathbb{R}^{N\times4096}$ be a sample of its inputs (the GELU outputs of $N$ activation tokens), $Y=AW^\top$ the output signal, $\tilde Y=A\tilde W^\top$ the compressed output and $E=\tilde Y-Y$ the error.
The second-moment matrices are $C=Y^\top Y/N=W\Sigma W^\top$ and $\tilde C=\tilde Y^\top\tilde Y/N=\tilde W\Sigma\tilde W^\top$ with $\Sigma=A^\top A/N$, so \emph{every measure below depends on the sample only through $\Sigma$}.
Write $C=\sum_i\lambda_i v_iv_i^\top$ with $\lambda_1\ge\lambda_2\ge\dots$; the \emph{span} of the output is the subspace of the eigenvectors $v_i$ with the largest eigenvalues $\lambda_i$: where the output has its energy. Four measures describe it. The \emph{spectrum angle} is the angle between the two spectra taken as vectors, $\theta(\tilde s_{1:k},\lambda_{1:k})$, with $\tilde s_1\ge\tilde s_2\ge\dots$ the eigenvalues of $\tilde C$ and $k=32$: the eigenvalues are ordered, so no eigenvectors have to be matched and the number does not depend on which channels hold the energy. It is zero when the compressed output keeps the shape of the spectrum, and it turns when energy is lost where the signal has most of it.
The \emph{spectrum distance} $d=\|\tilde s_{1:k}-\lambda_{1:k}\|/\|\lambda_{1:k}\|$ is a second form of the same comparison, which also counts a uniform change of scale.
The \emph{dominant gain} $\mu_1=v_1^\top\tilde Cv_1/\lambda_1$ is the variance kept along the dominant eigenvector; if $\tilde Y=\alpha Y+n$ along $v_1$, with $n$ the part not proportional to the signal, then $\mu_1=\alpha^2+\nu$ and the error energy along it is $(1-\alpha)^2+\nu$, so a small $\mu_1$ with a large error is a contraction. The \emph{in-span share} $\sum_{i\le k}v_i^\top\tilde Cv_i\big/\sum_{i\le k}\tilde s_i$ is the part of the compressed output that still lives where the signal lived (by Ky Fan's maximum principle the denominator is the most any $k$-dimensional subspace can hold). The \emph{error in the span} is the share of $\|E\|^2$ that falls along the top 8 eigenvectors of $Y$, together with the error relative to the signal along each eigenvector, $e_i=v_i^\top(E^\top E/N)v_i/\lambda_i$.
All four are eigenvalues or traces of $1024\times1024$ matrices; none needs a network run or a label. They avoid subspace angles on purpose: eigenvectors are poorly estimated from a small sample, and we saw the same difficulty
(the principal angle between the two top-8 subspaces is 17$^\circ$ for the rotation, whose error is $10^{-4}$ of the signal, because the cut splits near-degenerate eigenvectors).

\begin{table}[H]
\centering\small
\begin{tabular}{p{3.6cm}p{6.2cm}p{5.2cm}}
\toprule
measure & needs & sees\\
\midrule
weight angle & the matrix and its copy & the weight in the isotropic metric\\
branch angle (median over tokens) & the matrix, its copy, an input sample & the typical token\\
span measures & the matrix, its copy, the second moment of the inputs & the leading eigenvectors and the whole spectrum\\
stream angle & the network, compressed and float32 & what arrives from the other blocks\\
agreement, KL, swing & the whole network and a sample & the end result\\
\bottomrule
\end{tabular}
\caption{The measures compared. The span measures sit between the weight and the stream.}
\label{tab:measures}
\end{table}

\subsection{The span for the generalization of token samples}
\label{sec:smallsample}
The span carries what matters of a set of tokens, so from a small sample we can extrapolate it and rebuild a clone of the inputs of a block: tokens that satisfy the same basic requirements as the real ones.
From a small sample of real tokens $z$ at the input of the feed-forward part of a block\footnote{500 images, 98,500 tokens. The span measures of a block computed on these 500 images, on the 4,500 other images of sample 5000-995 and on all 5,000 agree to the digits we report (spectrum distance of the fitted table at block 11: 0.803, 0.804, 0.804). This is expected: we measure the error of a computation, which is mechanical, and not a classification, which changes label on a few images.} we estimate by a fixed rule the span of the standardized tokens and how the tokens load on it.
Most tokens form a \emph{bulk}, Gaussian with the mean and the covariance of the typical tokens (the covariance turns out to be optional, the mean is not). A few tokens are \emph{hot}: they load on one \emph{hot eigenvector} $u_1$, the first eigenvector of the second moment of the standardized tokens with the largest energy (the top 1\%), with its sign fixed positive.
The \emph{hot share} $p$ is the fraction of tokens whose projection on $u_1$ exceeds three standard deviations of the others' projections, and the \emph{amplitude} of the load is a log-normal fitted to the projections of the hot tokens.
A hot token is $z=c\sqrt n\,u_1+\sqrt{1-c^2}\,b$ with $b$ a bulk token orthogonalized to $u_1$ and scaled to norm $\sqrt n$ and $c$ the amplitude over $\sqrt n$, then standardized. The description is $u_1$ (about six channels), $p$, two amplitude numbers and the mean: about ten numbers.

%% file: sections/05_observations.tex
\section{Observations and findings}
\label{sec:obs}

{\small
\begin{longtable}{p{0.5cm}p{4.9cm}p{3.7cm}p{4.1cm}}
\caption{Observations, the test that could refute each, and the status.}\label{tab:obs}\\
\toprule
& observation & measurement & status\\
\midrule
\endfirsthead
\toprule
& observation & measurement & status\\
\midrule
\endhead
O1a & $(WT^{-1})_{ji}\sim N(0,\sigma_j^2)$
   & Kolmogorov--Smirnov
   & fc2 0/24 rejected, fc1 20/24 rejected\\
O1b & $b=4$: $\theta\big(WT^{-1},Q(WT^{-1})\big)=5.59^\circ$
   & weight angle
   & fc2: true (median 5.57$^\circ$, no token above 6.3$^\circ$); fc1: only the median (5.52$^\circ$, tokens up to 7.7$^\circ$)\\
O2 & The unrotated weight tokens have heavy tails, and the Lloyd--Max schemes (\SLMc{} and \FLMc{}) lose to \Grid{} in the tail.
   & 99th percentile of the weight angle, 8 bits
   & \FLMc{} 48/48, \SLMc{} 45/48\\
O3 & Spearman correlation of the stream angle after the last block, the logit angle and the KL divergence
   & rank correlation over the schemes and the widths
   & stream--logit 0.988, stream--KL 0.997, logit--KL 0.990; within one width 1.00 at 8, 6 and 4 bits, 0.80 at 2 bits\\
O4 & top-1 and the share of changed images, at 4 bits
   & 95\% intervals, per scheme and per sample
   & top-1: intervals overlap in 6/6 pairs of schemes on 5000-995 and on 50000-1000; changed share: disjoint in 5/6 and 6/6 pairs, the same on both samples for every scheme\\
O5 & rotation: independent errors, $\varphi_n\propto n^{1/2}$
   & exponent $p$ of $\varphi_n\propto n^{p}$
   & rotation $p=0.43$ to 0.53 at 8, 6, 4 and 2 bits; the crop schemes rise above it at 6 and 4 bits, \Grid{} not at 8 and 2 bits\\
O6 & $\varphi(x'_b)\gg\varphi_{\mathrm{FFN}}$, and $\varphi_{\mathrm{FFN}}$ does not depend on the input
   & median angle at the end of block 20: arriving, and added by the block with the clean and with the in-context input
   & arriving 12.1$^\circ$, added 2.1 to 2.9$^\circ$; the added error changes by at most 3\% with the input\\
O7 & \FLMc{} $<$ \Grid{} alone, $>$ in all 48 matrices
   & logit angle, 4 bits
   & $<$ in 4 of 4 blocks; $16.45^\circ>9.41^\circ$\\
O8 & the gap in $\varphi_{23}$ between all \FLMc{} and all \Grid{} ($6.9^\circ$) arises in fc2
   & \FLMc{} swapped into fc1 or fc2 of blocks 8--15 of the \Grid{} network
   & fc2 $+4.2^\circ$, fc1 $+0.2^\circ$\\
O9 & the sensitivity of a block in isolation at low bits locates the blocks that matter in context
   & overlap of the top 5 blocks, isolation and context
   & \SLMr{} 2 bits 4/5, \Grid{} 3 bits 5/5, \FLMc{} 8 bits 4/5; \FLMc{} and \SLMc{} 4 bits 1/5\\
O10 & from 500 images and their eigen factors we reproduce the span measures of the real inputs
   & Spearman correlation of the span measures, real against generated, over scheme-by-measure pairs
   & blocks 11--14: 0.98 to 1.00; block 17: 0.98 with the mean, 0.60 without\\
\bottomrule
\end{longtable}}

%% file: sections/06_results.tex
\section{Results}
\label{sec:results}

We follow the error from the weights to the classification and through the blocks of the network (the weights, classification, the error flow), then into single blocks (isolation and context, the span of fc2, its trigger, the input scheme), and end with mixed policies.

\subsection{The weights}
\label{sec:weights}
After a random-sign Hadamard rotation the coordinates of the weight tokens are Gaussian in fc2 and closer to Gaussian in fc1 (O1a); the unrotated tokens have heavy tails, the largest coordinate of a matrix being 10 to 37 sigma (O2).
For a Gaussian token a table with relative squared error $D$ (after the best scale of the compressed token) turns it by $\theta=\arcsin\sqrt D$, which is 5.59$^\circ$ at 4 bits (O1b). Tables~\ref{tab:wangle} and~\ref{tab:w8} give the weight angle at 4 and 8 bits.

\begin{table}[H]
\centering\small
\begin{tabular}{lcccc}
\toprule
4 bits, median angle ($^\circ$) & \SLMr & \Grid & \FLMc & \SLMc\\
\midrule
fc1 (median over 24 blocks) & 5.52 & 6.33 & 5.67 & 5.62\\
fc2 (median over 24 blocks) & 5.57 & 7.01 & 6.13 & 6.24\\
\bottomrule
\end{tabular}
\caption{Weight angle at 4 bits.}
\label{tab:wangle}
\end{table}

\begin{table}[H]
\centering
\begin{tabular}{lcccc}
\toprule
8 bits, over the 48 matrices & median & 99th pct & max & smallest norm ratio\\
\midrule
\Grid & 0.52 & 1.41 & 3.12 & 0.997\\
\FLMc & 0.50 & 2.50 & 10.96 & 0.881\\
\SLMc & 0.38 & 8.50 & 28.42 & 0.680\\
\SLMr & 0.36 & 0.42 & 0.74 & 0.999\\
\bottomrule
\end{tabular}
\caption{Weight angle (degrees) at 8 bits: median over the matrices of the per-matrix median, 99th percentile and maximum over tokens, and the smallest ratio $\|\tilde w\|/\|w\|$.}
\label{tab:w8}
\end{table}

\subsection{Classification}
\label{sec:class}
Table~\ref{tab:sweep} gives the sweep on 5,000 images and Table~\ref{tab:val50k} the result at 4 bits on 50,000. The rotation is the closest to float32 at every width; the agreement falls gently from 8 to 4 bits and steeply below 3 bits, and the logit angle of the rotation grows by a factor of 1.98 to 1.99 per bit from 8 to 3 bits, as a weight distortion that falls by four per bit would give (O3).
At 4 bits \FLMc{} is below \Grid{} by $-1.23$ [$-1.40$, $-1.07$] points of agreement and $-0.30$ [$-0.44$, $-0.15$] of top-1 (paired, 50,000 images). The intervals of top-1 overlap for every pair of schemes on both samples, while the share of images that change class is 1.6, 3.1, 4.4 and 5.2 to 5.6\% and is the same on both (O4).

\begin{table}[H]
\centering\footnotesize
\begin{tabular}{llcccccc}
\toprule
 & & 8 & 6 & 5 & 4 & 3 & 2 bits\\
\midrule
agreement (\%) & \SLMr & 99.92 & 99.56 & 98.92 & 98.44 & 96.70 & 92.22\\
 & \Grid & 99.78 & 99.28 & 98.24 & 96.90 & 94.18 & 82.34\\
 & \FLMc & 99.40 & 98.90 & 97.82 & 95.64 & 93.34 & 85.68\\
 & \SLMc & 96.78 & 95.82 & 95.62 & 94.76 & 92.56 & 83.52\\
logit angle ($^\circ$) & \SLMr & 0.23 & 0.90 & 1.79 & 3.54 & 7.02 & 15.37\\
 & \Grid & 0.45 & 2.09 & 4.60 & 9.41 & 15.12 & 28.46\\
 & \FLMc & 1.89 & 2.78 & 5.87 & 16.45 & 21.75 & 30.79\\
 & \SLMc & 11.48 & 14.50 & 16.74 & 19.24 & 23.29 & 32.04\\
mean KL & \SLMr & 0.0000 & 0.0003 & 0.0011 & 0.0044 & 0.0182 & 0.1087\\
 & \Grid & 0.0001 & 0.0011 & 0.0074 & 0.0267 & 0.0698 & 0.4574\\
 & \FLMc & 0.0007 & 0.0015 & 0.0076 & 0.0559 & 0.1127 & 0.4581\\
 & \SLMc & 0.0326 & 0.0448 & 0.0573 & 0.0772 & 0.1377 & 0.5404\\
\bottomrule
\end{tabular}
\caption{Sweep on 5,000 images, all 48 FFN matrices compressed.}
\label{tab:sweep}
\end{table}

\begin{table}[H]
\centering\footnotesize
\begin{tabular}{lcccccc}
\toprule
 & agreement (\%) & top-1 & top-5 & changed / broken / fixed & logit angle & KL\\
\midrule
float32 & -- & 85.84 & 97.82 & -- & -- & --\\
\SLMr & 98.40 [98.29, 98.51] & 85.69 & 97.83 & 802 / 320 / 244 & 3.54 & 0.0041\\
\Grid & 96.84 [96.69, 96.99] & 85.59 & 97.73 & 1580 / 623 / 501 & 9.35 & 0.0264\\
\FLMc & 95.61 [95.44, 95.80] & 85.29 & 97.64 & 2193 / 923 / 650 & 16.49 & 0.0556\\
\SLMc & 94.42 [94.22, 94.62] & 85.06 & 97.56 & 2792 / 1200 / 812 & 19.27 & 0.0775\\
\bottomrule
\end{tabular}
\caption{4 bits, 50,000 validation images, all 48 matrices.}
\label{tab:val50k}
\end{table}

\subsection{The error flow}
\label{sec:flow}
\begin{figure}[H]
\centering
\includegraphics[width=0.95\linewidth]{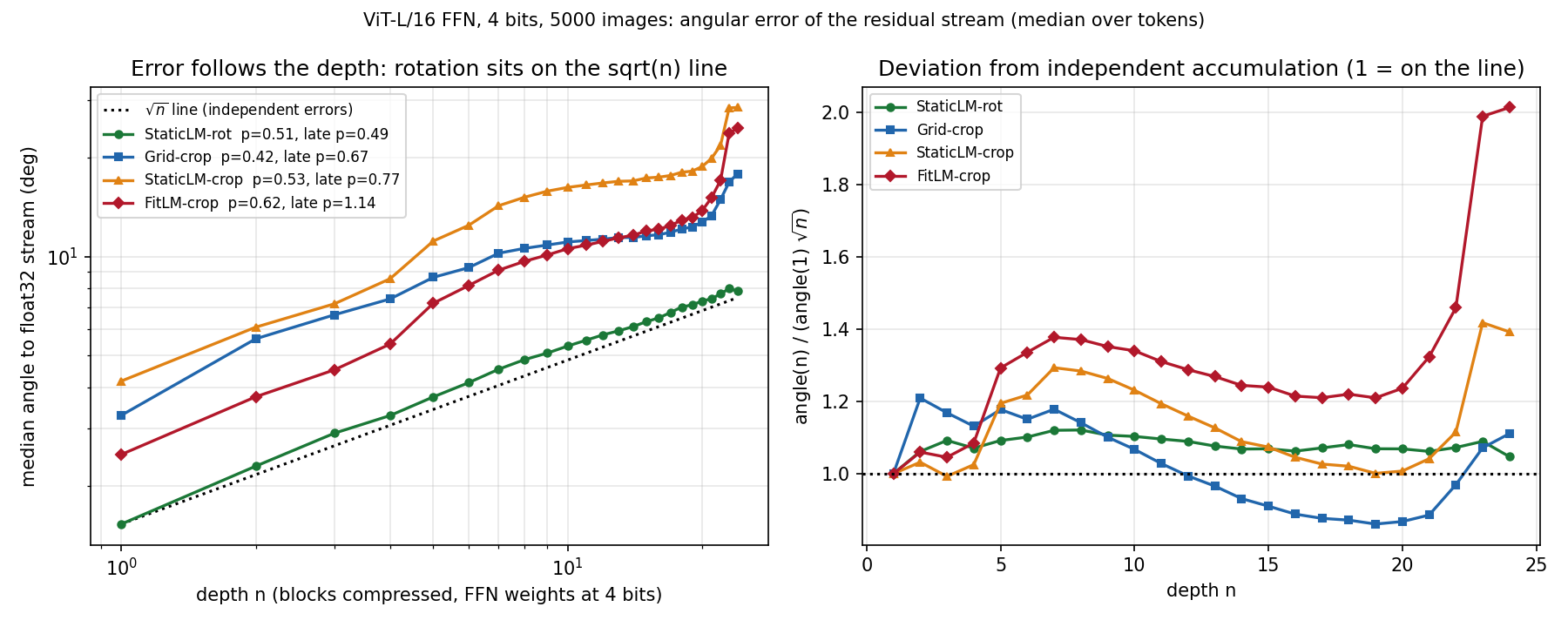}
\caption{Median angle of the residual stream against the number of blocks, 4 bits, 5,000 images. Left: log--log with the $\sqrt n$ line. Right: the angle divided by $\varphi_1\sqrt n$.}
\label{fig:flow4}
\end{figure}

Figure~\ref{fig:flow4} shows the median stream angle after each block at 4 bits and Figure~\ref{fig:specflow4} the spectrum angle of the stream; the other widths are in Appendix~\ref{sec:appflow}. Table~\ref{tab:exp} gives the growth exponents.
The rotation follows $n^{1/2}$ at every width (O5) and the crop schemes rise above it in the later blocks. The spectrum angle is not cumulative: it rises where a scheme loses energy (after the 12th to the 14th block, that is blocks 11 to 13, for the Lloyd--Max schemes at 4 bits) and falls back in the later blocks, while the stream angle keeps growing.

\begin{figure}[H]
\centering
\includegraphics[width=0.95\linewidth]{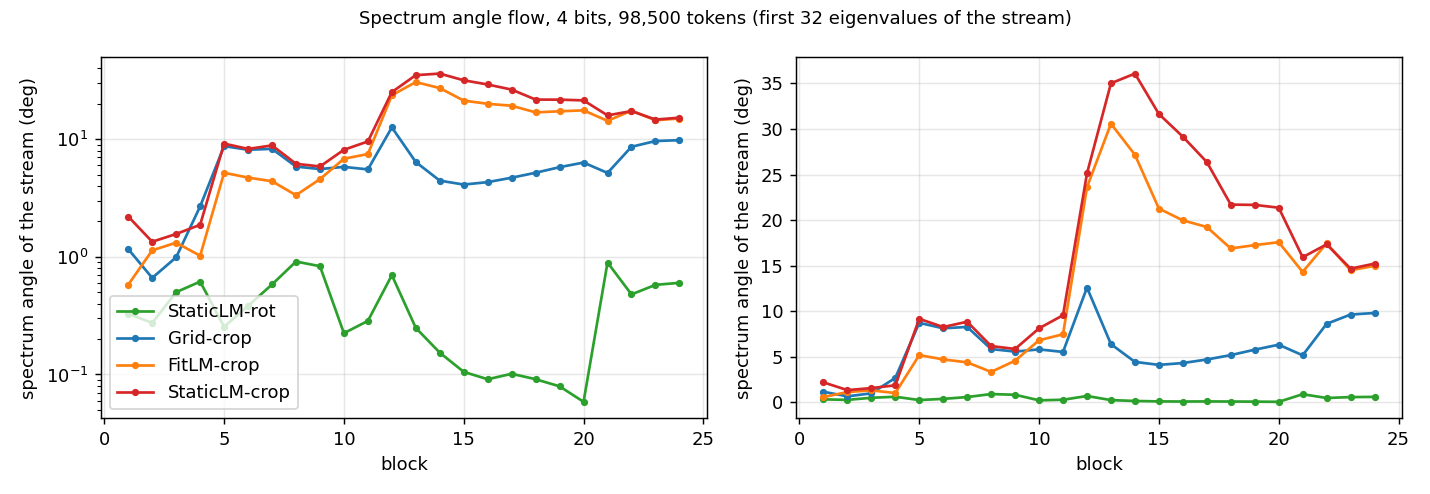}
\caption{Spectrum angle of the residual stream against the number of blocks, 4 bits, 500-401 (98,500 tokens): the angle between the first 32 eigenvalues of the second moment of the stream, compressed against float32. Left: log scale. Right: linear.}
\label{fig:specflow4}
\end{figure}

\begin{table}[H]
\centering\small
\begin{tabular}{llcccc}
\toprule
 & & 8 & 6 & 4 & 2 bits\\
\midrule
exponent $p$ (all / late) & \SLMr & 0.53 / 0.54 & 0.53 / 0.53 & 0.51 / 0.49 & 0.43 / 0.29\\
 & \Grid & 0.48 / 0.43 & 0.51 / 0.73 & 0.42 / 0.67 & 0.37 / 0.13\\
 & \FLMc & 0.75 / 1.64 & 0.61 / 1.14 & 0.62 / 1.14 & 0.46 / 0.49\\
 & \SLMc & 0.63 / 1.34 & 0.61 / 1.11 & 0.53 / 0.77 & 0.42 / 0.45\\
angle after block 24 ($^\circ$) & \SLMr & 0.52 & 2.02 & 7.85 & 32.27\\
 & \Grid & 0.94 & 4.63 & 17.88 & 52.50\\
 & \FLMc & 3.99 & 5.30 & 24.62 & 51.34\\
 & \SLMc & 19.10 & 22.13 & 28.50 & 53.04\\
\bottomrule
\end{tabular}
\caption{Growth exponents (over all 24 blocks, and over blocks 13--24) and the final stream angle, 5,000 images. }
\label{tab:exp}
\end{table}

\subsection{Isolation and context}
\label{sec:iso}
Compressed alone, the fitted table is as good as or better than the grid in blocks 0, 1, 20 and 21; in all 48 matrices it is worse (O7). The error that arrives at block 20 is four to five times the error the block adds, and the error the block adds hardly depends on its input (Table~\ref{tab:ctx}, O6).
Swapping the fitted table into the grid network puts the gap in fc2 (Table~\ref{tab:swap}, O8). The sensitivity of a block in isolation locates the blocks that matter in context for a mild error, and not for the table schemes at 4 bits (O9): isolation points to blocks 23, 22, 11, 21 and 0, context to blocks 11, 12, 13, 17 and 14.

\begin{table}[H]
\centering
\begin{tabular}{lccc}
\toprule
median angle at the end of block 20 ($^\circ$) & \SLMr & \Grid & \FLMc\\
\midrule
local error, clean input & 2.19 & 2.88 & 2.45\\
local error, in-context input & 2.11 & 2.90 & 2.49\\
total error (arriving + local) & 12.33 & 12.63 & 12.57\\
\bottomrule
\end{tabular}
\caption{Block 20 at 4 bits, 500 images. The error that arrives from the blocks before is 12.14$^\circ$.}
\label{tab:ctx}
\end{table}

\begin{table}[H]
\centering
\begin{tabular}{lc}
\toprule
network (4 bits, 500 images) & angle after block 23 ($^\circ$)\\
\midrule
all \Grid & 17.87\\
all \FLMc & 24.75\\
\Grid with \FLMc{} in blocks 0--7 / 8--15 / 16--21 & 17.88 / 22.21 / 18.41\\
\Grid with \FLMc{} in blocks 22--23 & 18.83\\
\Grid with \FLMc{} in fc1 / fc2 of blocks 8--15 & 18.05 / 22.06\\
\FLMc{} with \Grid{} in blocks 0--7 / 8--15 / 16--21 & 24.73 / 19.57 / 22.86\\
\FLMc{} with \Grid{} in blocks 22--23 & 23.77\\
\FLMc{} with \Grid{} in fc1 / fc2 of blocks 8--15 & 24.66 / 19.67\\
\bottomrule
\end{tabular}
\caption{Swap tests: median stream angle after the last block.}
\label{tab:swap}
\end{table}

\subsection{The error of fc2 in the span of its output}
\label{sec:spanresults}
Table~\ref{tab:card} gives the span measures of Section~\ref{sec:spanmeasures} for fc2 of blocks 11 and 17, compressed alone, at 4 bits.
At block 11 the fitted table loses the dominant eigenvector of the output ($\mu_1=0.09$), made of the stream channels 681 and 724, rows of W2 with coordinates of 22 to 27 sigma: the weight angle of these rows is 1.5 times the grid's and the output error 400 times. The spectrum angle orders the schemes as the end of the flow does, the median branch angle does not.
At block 17 both crop schemes lose the dominant eigenvector (spectrum angle 25.6 and 29.8$^\circ$) and the rotation does not (0.11$^\circ$).

\begin{table}[H]
\centering\footnotesize\setlength{\tabcolsep}{3pt}
\begin{tabular}{p{3.3cm}cccccc}
\toprule
 & \multicolumn{3}{c}{block 11} & \multicolumn{3}{c}{block 17}\\
 & \FLMc & \Grid & \SLMr & \FLMc & \Grid & \SLMr\\
\midrule
spectrum angle ($^\circ$) & 19.77 & 4.46 & 0.02 & 25.60 & 29.84 & 0.11\\
spectrum distance & 0.803 & 0.296 & 0.016 & 0.827 & 0.818 & 0.011\\
dominant gain $\mu_1$ & 0.090 & 0.647 & 0.983 & 0.047 & 0.087 & 0.986\\
in-span share (32) & 0.919 & 0.951 & 0.999 & 0.957 & 0.961 & 0.997\\
error in the top-8 eigenvectors (\%) & 93.2 & 44.8 & 3.0 & 91.6 & 88.6 & 1.1\\
error over signal, top 8 & 0.429 & 0.038 & 0.0001 & 0.472 & 0.382 & 0.0002\\
error on the top 1\% of tokens (\%) & 68.1 & 62.2 & 6.3 & 52.6 & 51.8 & 5.7\\
branch angle, median ($^\circ$) & 6.80 & 8.29 & 5.75 & 5.95 & 6.74 & 5.69\\
final angle, isolation ($^\circ$) & 3.43 & 1.62 & 0.97 & 1.42 & 1.47 & 1.06\\
\bottomrule
\end{tabular}
\caption{fc2 of one block compressed alone at 4 bits, 500 images. The span measures of Section~\ref{sec:spanmeasures}, the concentration of the error on tokens, the median branch angle (the angle between the output of fc2 and its compressed counterpart, per token) and the final stream angle.}
\label{tab:card}
\end{table}

\subsection{What triggers the error: the input}
\label{sec:trigger}
Gaussian tokens do not trigger the loss: with the real mean and covariance they give a spectrum distance of 0.03 for the fitted table at block 11, against 0.80 for the real input. Independent hidden units with the real energy of each unit reproduce most of it (0.53), and the same energies permuted over the units remove it (0.02): it is the alignment of the input energy with the outlier columns of the heavy rows that triggers the error.
At block 11 that energy comes from about 1\% of the tokens, which lie along one eigenvector $u_1$ of six channels (724, 681, 319, 612, 427, 131).

\subsection{The input scheme across blocks}
\label{sec:schemeblocks}
Table~\ref{tab:blocks} applies the description of Section~\ref{sec:smallsample}, estimated on 500 images of each block, to seven blocks and judges it on all 5,000 images, 4,500 of which were not used for the estimation.
At blocks 11 to 14 it reproduces the span measures and the order of the schemes (Spearman 0.98 to 1.00); block 17 needs the mean of the tokens (0.60 without, 0.98 with); the quiet blocks 5 and 8 have no hot population, and the description reproduces their lack of failure (spectrum distance of the fitted table 0.023 and 0.097 against the real 0.026 and 0.031).
Where the local measures cannot tell the two crop schemes apart (blocks 14 and 17), the description gives the same answer.

\begin{table}[H]
\centering\footnotesize\setlength{\tabcolsep}{4pt}
\begin{tabular}{rccccc}
\toprule
block & $p$ (\%) & ch. 60\% & spectrum dist. & Spearman & table $-$ grid\\
 & & of $u_1$ & real / scheme & 15 pairs & real / scheme\\
\midrule
5 & 18.7$^*$ & 63 & 0.026 / 0.023 & 0.957 & $-0.009$ / $-0.006$\\
8 & 97.6$^*$ & 37 & 0.031 / 0.097 & 0.814 & $-0.002$ / $+0.017$\\
11 & 5.1 & 4 & 0.804 / 0.750 & 1.000 & 0.509 / 0.465\\
12 & 5.1 & 3 & 0.877 / 0.760 & 1.000 & 0.547 / 0.443\\
13 & 4.6 & 3 & 0.843 / 0.662 & 0.975 & 0.347 / 0.189\\
14 & 4.3 & 3 & 0.811 / 0.600 & 0.993 & 0.050 / 0.012\\
17 & 4.6 & 3 & 0.834 / 0.929 & 0.979 & 0.009 / 0.044\\
\bottomrule
\end{tabular}
\caption{The input scheme (with the mean) estimated on 500 images and compared with the real 5,000 images, fc2 at 4 bits: the hot share $p$ and the channels carrying 60\% of $u_1$; the spectrum distance of the fitted table, real and generated; the Spearman correlation over the scheme-by-measure pairs; and the spectrum distance of the fitted table minus that of the grid, real and generated.
$^*$ The quiet blocks have no hot population, so the three-sigma rule is degenerate.}
\label{tab:blocks}
\end{table}

\subsection{Mixed policies}
Rotating a few matrices does not recover the gap. With the fitted table elsewhere, rotating block 0 and fc1 of block 1 (3 matrices), 4 matrices, or the 12 most sensitive matrices gives 95.94, 95.82 and 95.76\% agreement (5,000 images), against 95.64\% for none and 98.44\% for all.
On a grid base, rotating blocks 0--1 gains 0.11 [0.01, 0.20] points of agreement on 50,000 images, and the fitted table in blocks 20--21 changes the agreement by $-0.04$ [$-0.14$, $+0.06$] while the mean KL rises by 0.0048.

%% file: sections/07_conclusion.tex
\section{Conclusion}
\label{sec:conclusion}

The angle between the compressed and the float32 residual stream, followed through the blocks, ranks the schemes as the classifier does and separates what top-1 and top-5 do not. The spectrum of the output of a matrix gives a measure that depends on the context, and a description of a few numbers estimated on 500 images generates inputs on which it can be studied.
Together they show the two regimes of the error (independent accumulation with the rotation, amplification with the crop schemes) and where a scheme loses: in fc2 of a few blocks, through the dominant eigenvector of its output.

\paragraph{Scope.} One model, the FFN weights, one random sign seed, and float32 evaluation of the compressed weights, so that we measure the representation and not a kernel and report no run time, including the cost of the rotation. The span analysis is at 4 bits and the flow curves are medians over tokens.

%% file: sections/08_appendix.tex
\appendix
\section{The error flow at 8, 6 and 2 bits}
\label{sec:appflow}

Figure~\ref{fig:flow4} for the other widths: the median angle of the residual stream against the number of blocks, 5,000 images. Left: log--log with the $\sqrt n$ line. Right: the angle divided by $\varphi_1\sqrt n$.

\begin{figure}[H]
\centering
\includegraphics[width=0.95\linewidth]{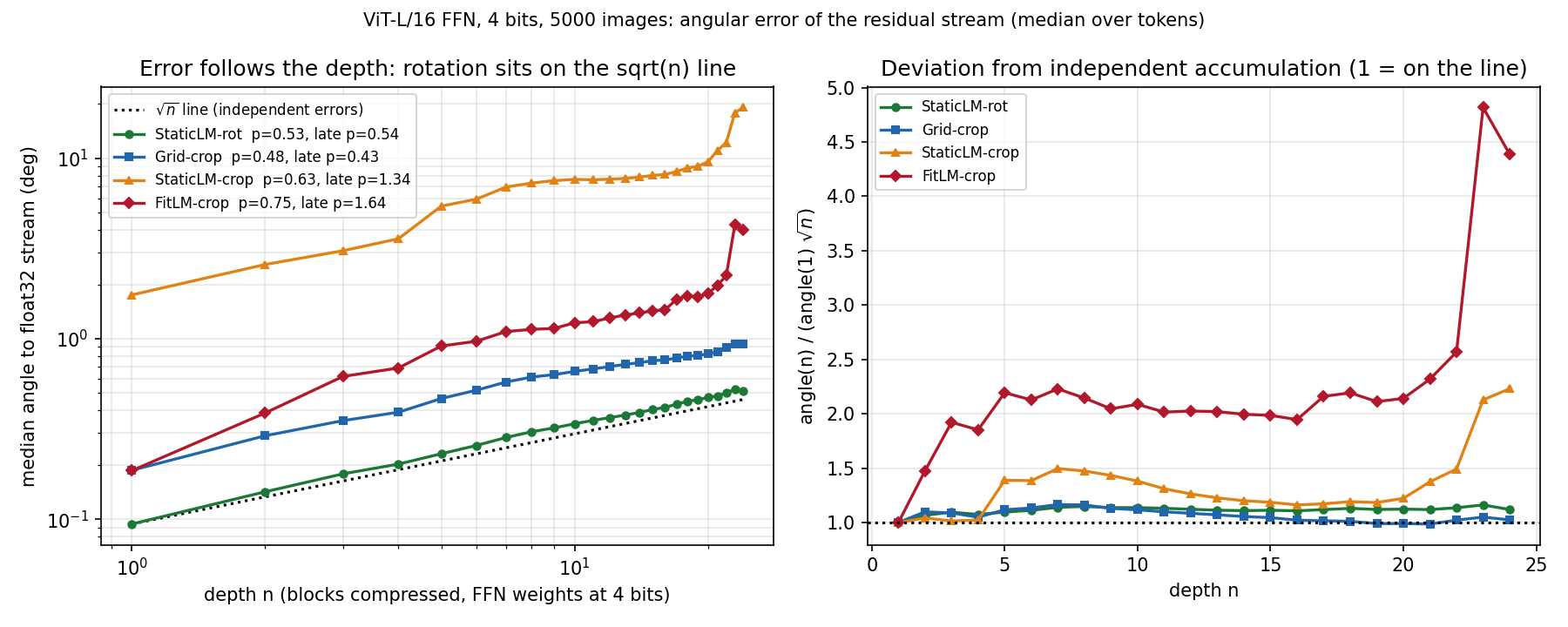}
\caption{Error flow, 8 bits.}
\label{fig:flow8}
\end{figure}

\begin{figure}[H]
\centering
\includegraphics[width=0.95\linewidth]{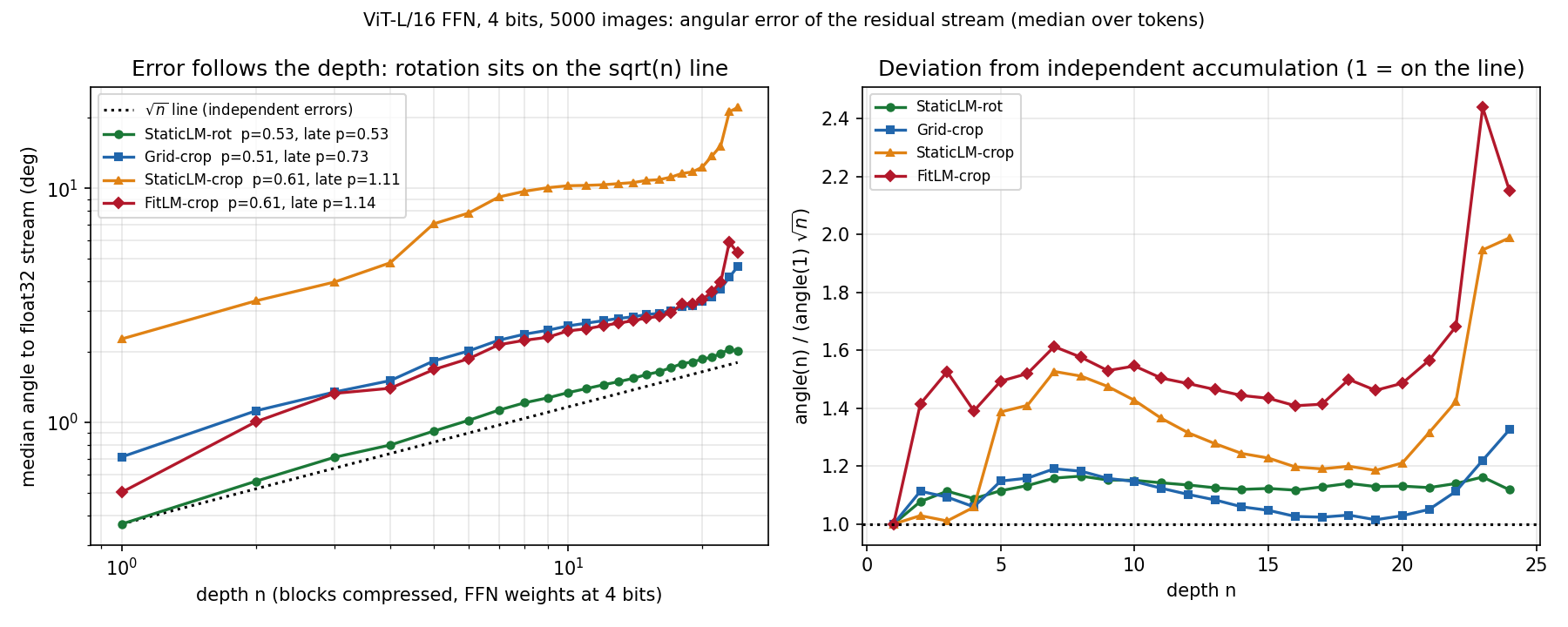}
\caption{Error flow, 6 bits.}
\label{fig:flow6}
\end{figure}

\begin{figure}[H]
\centering
\includegraphics[width=0.95\linewidth]{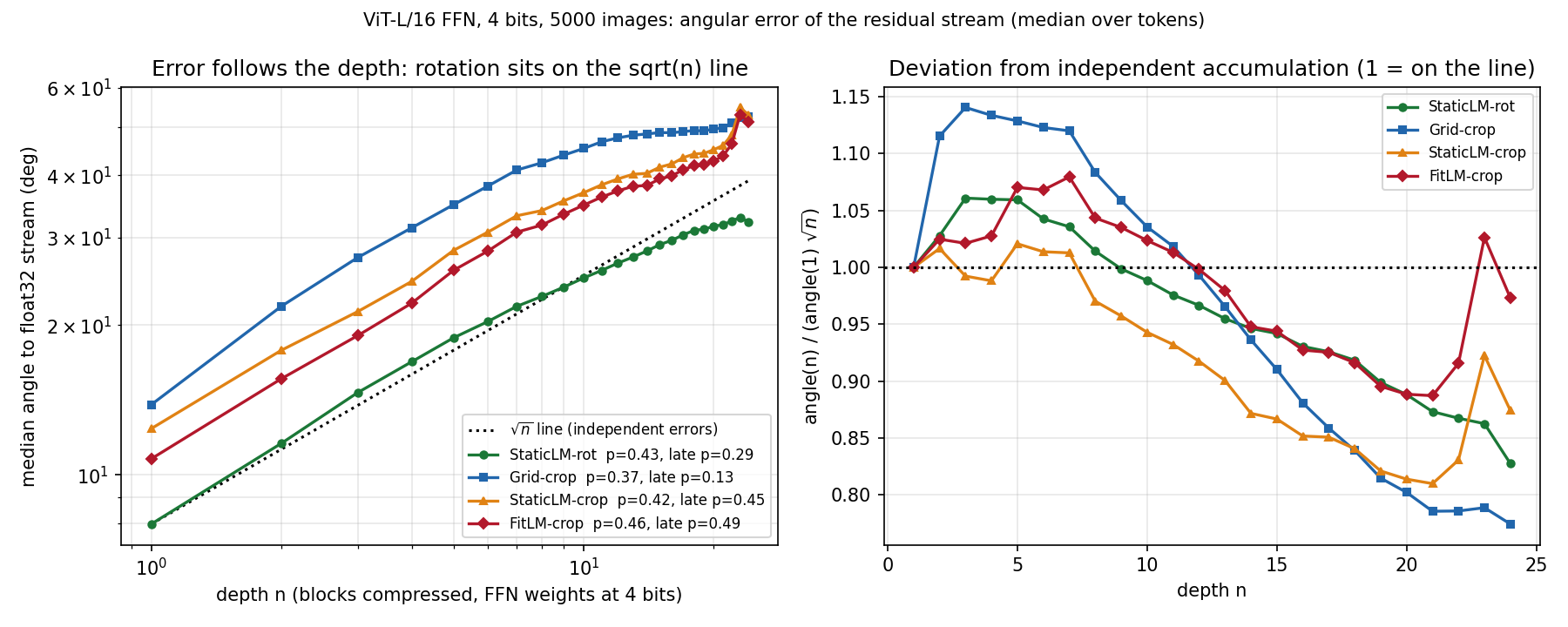}
\caption{Error flow, 2 bits.}
\label{fig:flow2}
\end{figure}

\section{The spectrum angle flow at 8, 6 and 2 bits}
\label{sec:appspec}

\begin{figure}[H]
\centering
\includegraphics[width=0.95\linewidth]{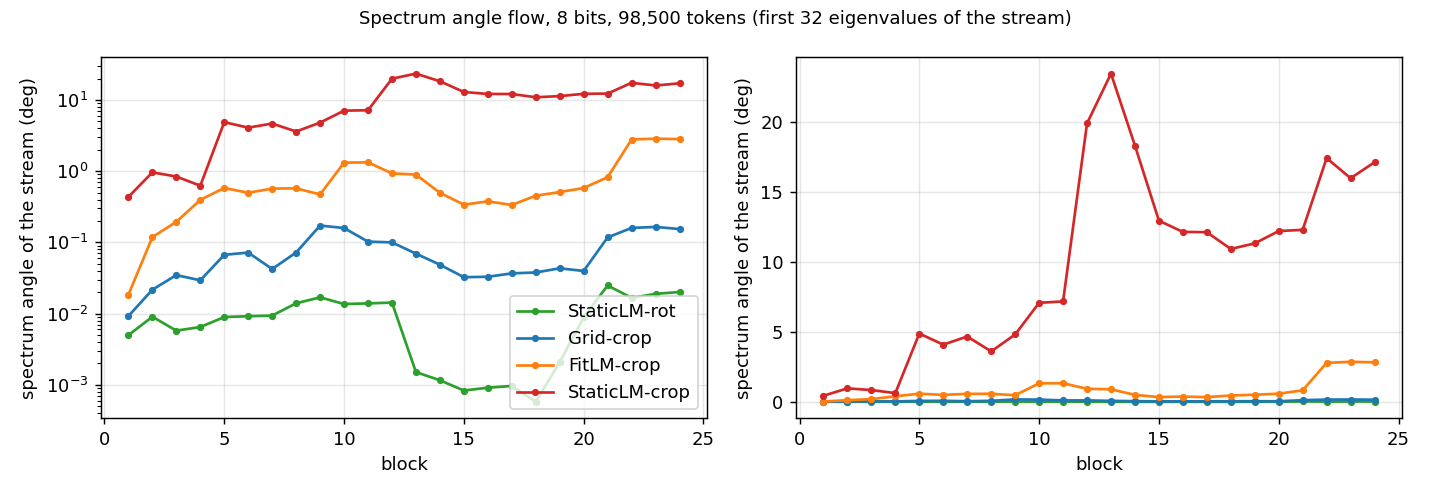}
\caption{Spectrum angle flow, 8 bits (as Figure~\ref{fig:specflow4}).}
\end{figure}

\begin{figure}[H]
\centering
\includegraphics[width=0.95\linewidth]{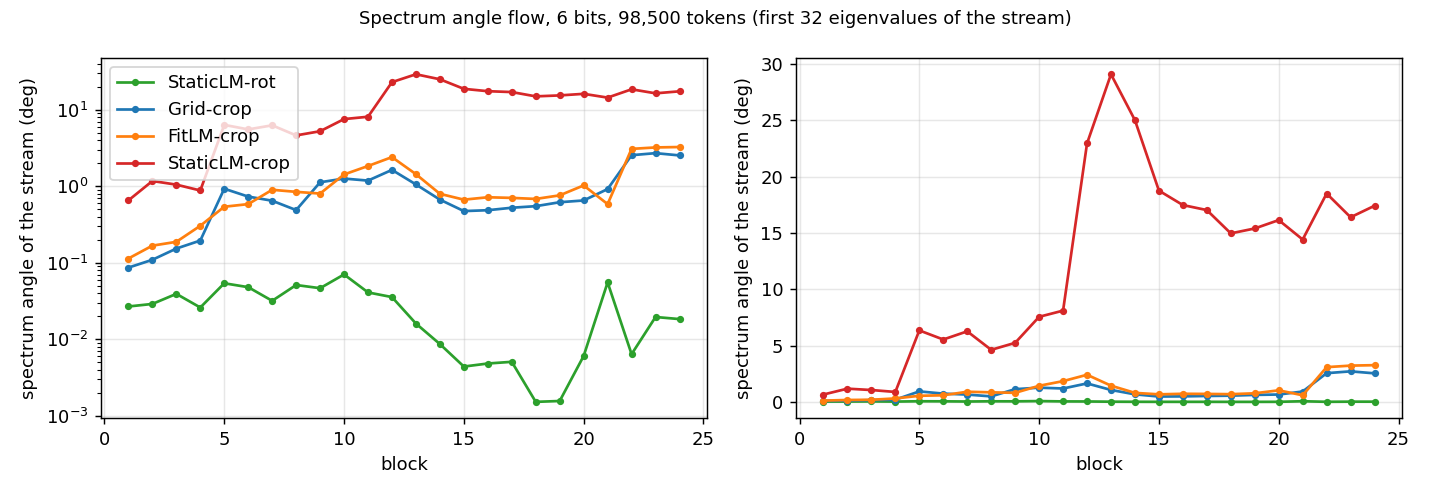}
\caption{Spectrum angle flow, 6 bits.}
\end{figure}

\begin{figure}[H]
\centering
\includegraphics[width=0.95\linewidth]{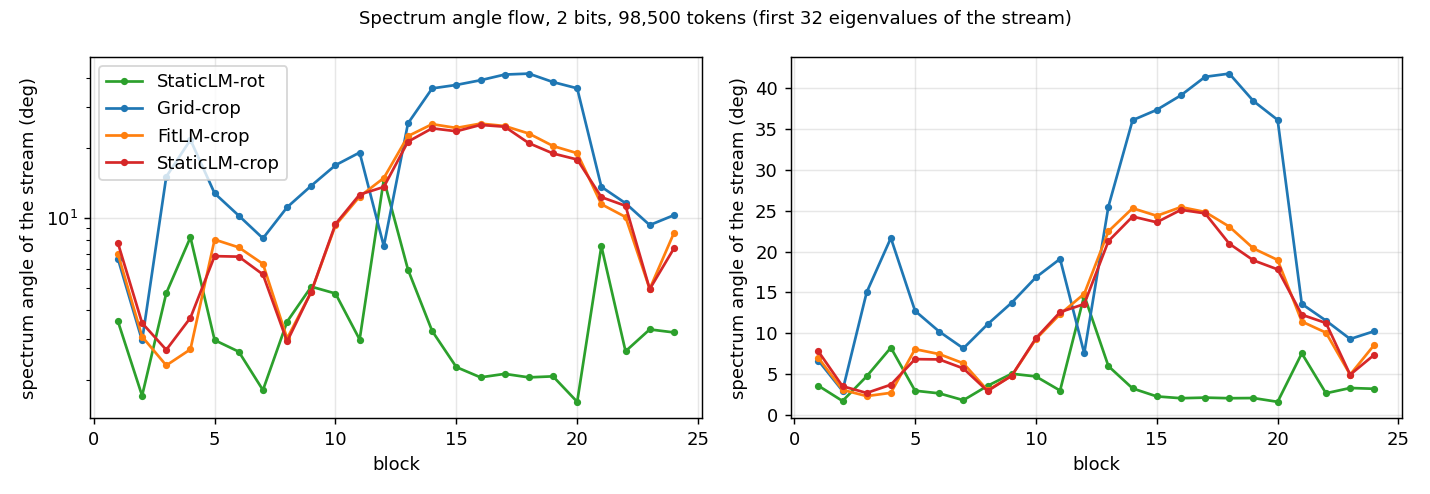}
\caption{Spectrum angle flow, 2 bits.}
\end{figure}